\documentclass[journal,compsoc]{IEEEtran}

\usepackage{silence}
\usepackage{xcolor}
\usepackage[normalem]{ulem}

\usepackage[pdftex]{graphicx}
\usepackage{microtype}
\usepackage{amsmath,amssymb}
\usepackage[table,xcdraw]{xcolor}
\usepackage{array}
\usepackage{cite}
\usepackage{multirow}
\usepackage{dirtytalk}
\usepackage{hyperref}
\usepackage{url}
\usepackage{xspace}
\usepackage{booktabs}
\usepackage{orcidlink}
\usepackage{algorithm}
\usepackage{algpseudocode}

\newcommand{\orcidID}[1]{\orcidlink{#1}}

\title{Contour-Guided Query-Based Feature Fusion for Boundary-Aware and Generalizable Cardiac Ultrasound Segmentation}

\author{%
Zahid~Ullah\,\orcidID{0000-0002-0184-7620},
Sieun~Choi\,\orcidID{0009-0006-6623-4645},
and~Jihie~Kim\,\orcidID{0000-0003-2358-4021}%
\IEEEcompsocitemizethanks{%
\IEEEcompsocthanksitem
Zahid Ullah, Sieun Choi, and Jihie Kim are with the Department of Computer Science and Artificial Intelligence, Dongguk University, Seoul 04620, Republic of Korea (e-mail: zahid1989@dongguk.edu; sieunchoi@dgu.ac.kr; jihie.kim@dgu.edu).

\IEEEcompsocthanksitem
Corresponding author: Jihie Kim (e-mail: jihie.kim@dgu.edu).
}%
}

\begin{document}
\maketitle

\begin{abstract}
 Accurate cardiac ultrasound segmentation is critical for reliable assessment of ventricular function in intelligent healthcare systems. However, echocardiographic images are inherently challenging due to low contrast, speckle noise, irregular anatomical boundaries, and significant domain shift across acquisition devices and patient populations. Existing methods, primarily driven by appearance-based learning, often struggle to maintain boundary precision and structural consistency under these conditions. To address these limitations, we propose a Contour-Guided Query Refinement Network (CGQR-Net) for boundary-aware cardiac ultrasound segmentation. The proposed framework performs effective information fusion by integrating multi-resolution feature representations with contour-derived structural priors. Specifically, a High-Resolution Network (HRNet) backbone preserves high-resolution spatial information while capturing multi-scale context. A coarse segmentation is first generated, from which anatomical contours are extracted and encoded into learnable query embeddings. These contour-guided queries interact with fused feature maps through cross-attention, enabling structure-aware refinement that enhances boundary delineation and suppresses noise-induced artifacts. In addition, a dual-head supervision strategy jointly optimizes segmentation and boundary predictions to enforce structural consistency. The proposed method is evaluated on the Cardiac Acquisitions for Multi-structure Ultrasound Segmentation (CAMUS) dataset and further validated on the CardiacNet dataset to assess cross-dataset generalization. Experimental results demonstrate that CGQR-Net achieves superior segmentation accuracy, improved boundary precision, and strong robustness across different imaging conditions. These findings highlight the effectiveness of integrating contour-level structural information with feature-level representations, providing a robust and generalizable solution for cardiac ultrasound segmentation in real-world clinical and consumer healthcare applications.
\end{abstract}

\begin{IEEEkeywords}
            Medical imaging in consumer devices, echocardiographic segmentation, contour-guided refinement, attention mechanisms, boundary-aware learning, intelligent healthcare systems, domain generalization.
\end{IEEEkeywords}

\section{Introduction}
\label{intro}

\begin{figure}[!ht]
     \centering
     \includegraphics[width=\columnwidth]{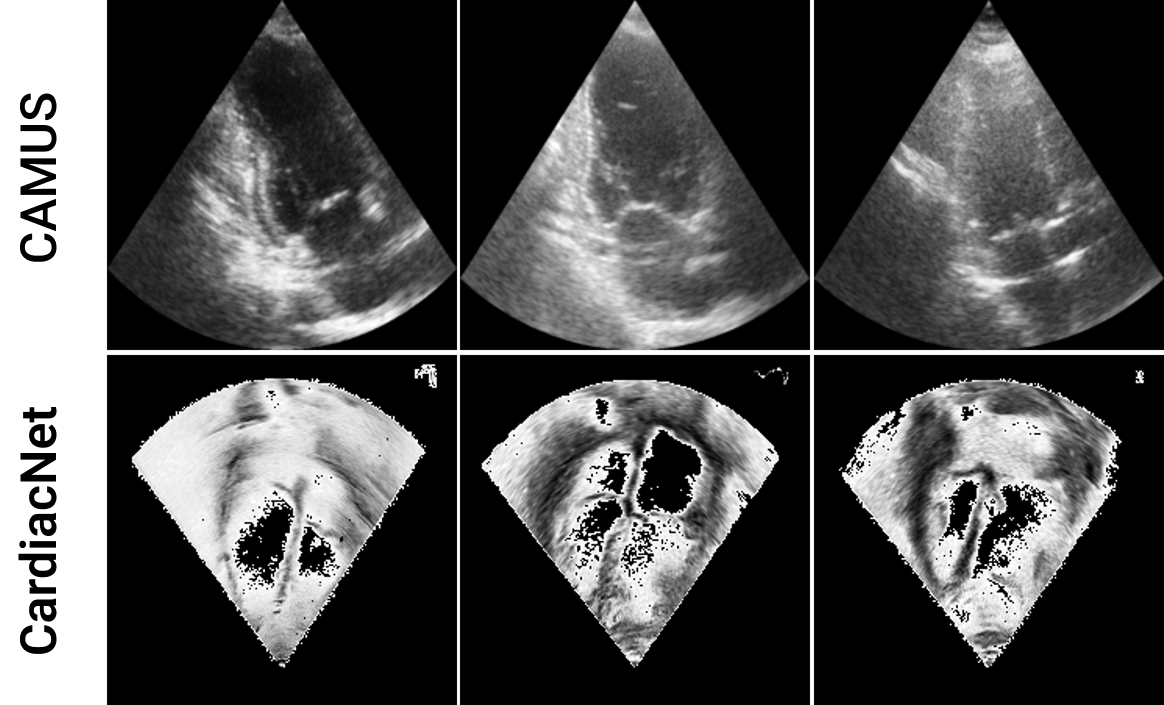}
     \caption{Sample echocardiographic images from the CAMUS (top row) and CardiacNet (bottom row) datasets, illustrating differences in image quality, noise levels, and anatomical variability. The CAMUS dataset contains relatively clean echocardiographic images with well-defined cardiac structures, while the CardiacNet dataset exhibits significantly higher variability, including severe speckle noise, low contrast, and irregular anatomical boundaries.}
     \label{datasets_sample}
\end{figure}

\begin{figure*}[!t] 
\centering 
\includegraphics[width=1\textwidth]{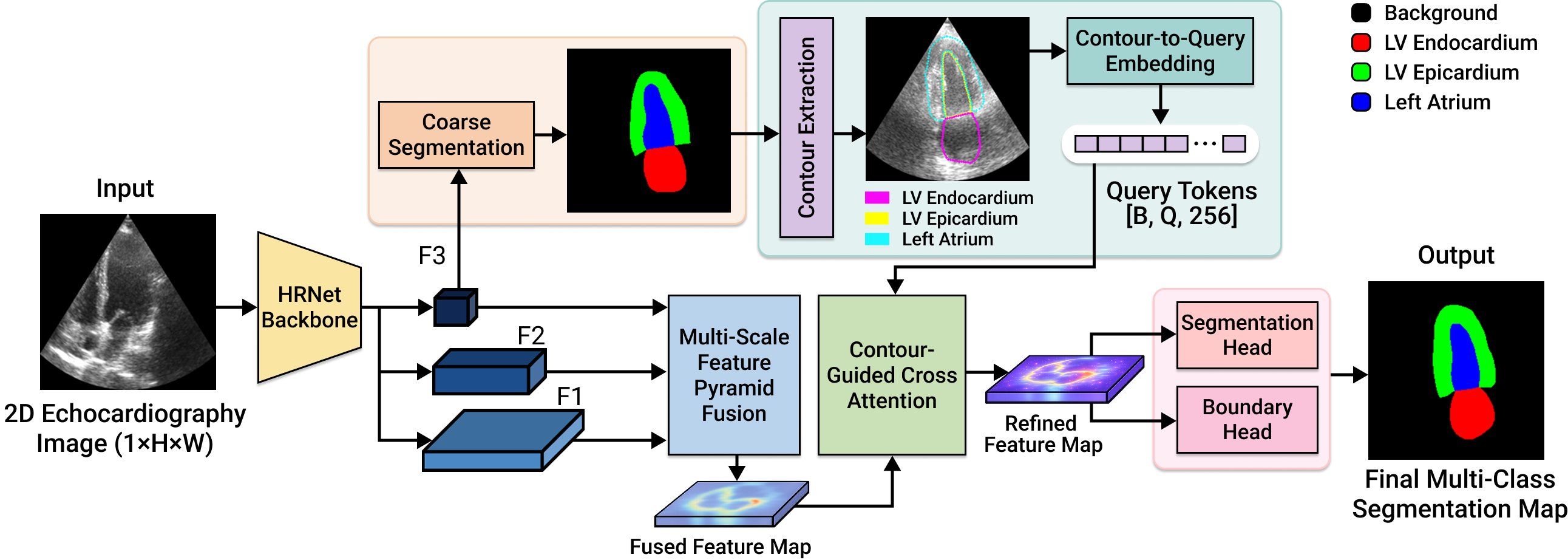} 
\caption{Architecture of the proposed CGQR-Net. HRNet extracts multi-resolution features from the input echocardiography image, and a coarse segmentation head provides an initial structural prediction. Contours extracted from the coarse mask are converted into query embeddings and used to refine fused multi-scale features through cross-attention. The refined representation is then passed to segmentation and boundary heads to produce the final boundary-aware multi-class segmentation.} 
\label{network} 
\end{figure*} 

\IEEEPARstart{C}{a}rdiac ultrasound, commonly known as echocardiography, is one of the most widely used imaging modalities for assessing cardiac structure and function \cite{lang2015recommendations}. It is non-invasive, cost-effective, portable, and capable of real-time visualization of cardiac motion. Clinical measurements such as left ventricular ejection fraction, ventricular volume, myocardial thickness, and atrial size critically depend on accurate delineation of cardiac structures. In particular, precise segmentation of the left ventricular (LV) endocardium, LV epicardium, and left atrium (LA) is essential for reliable assessment of cardiac function and diagnosis of cardiovascular diseases \cite{nagueh2016recommendations}.

Manual delineation of these structures is labor-intensive and subject to significant inter- and intra-observer variability, limiting reproducibility in clinical practice. Consequently, automatic cardiac ultrasound segmentation \cite{zhang2018fully} has attracted substantial attention, aiming to provide consistent and efficient analysis for intelligent healthcare systems. However, achieving accurate and reliable segmentation remains challenging due to the intrinsic characteristics of ultrasound imaging.

Echocardiographic images \cite{noble2006ultrasound} often exhibit low contrast, weak and ambiguous boundaries, and severe speckle noise. The intensity distributions of adjacent tissues, such as myocardium and blood pool, frequently overlap, making boundary localization difficult \cite{camus}. In addition, cardiac structures undergo complex non-linear deformations across the cardiac cycle, resulting in high anatomical variability across subjects and pathological conditions \cite{ouyang2020video}. These factors make it difficult for models \cite{oktay2018attention,zhou2018unet++} to simultaneously capture fine boundary details and global structural consistency. Furthermore, domain shift \cite{guan2021domain,perone2019unsupervised} caused by differences in acquisition devices, imaging protocols, and patient populations often leads to significant performance degradation when models are applied to unseen datasets.

Recent advances in deep learning, particularly convolutional neural networks (CNNs) \cite{ronneberger2015u}, have significantly improved medical image segmentation performance. Architectures such as U-Net \cite{ronneberger2015u} and High-Resolution Network (HRNet) \cite{wang2020deep} leverage multi-scale feature extraction to capture both local and contextual information. However, these models are primarily driven by pixel-wise supervision and local receptive fields \cite{long2015fully}. As a result, they often produce overly smooth or irregular boundaries, especially in low-contrast regions, and lack explicit mechanisms to enforce anatomical consistency \cite{chen2017deeplab}.

Transformer-based approaches \cite{dosovitskiy2020image,vaswani2017attention} have been introduced to address the limitation of local modeling by capturing long-range dependencies through self-attention. While these models improve global context representation, they typically operate on feature tokens without explicitly incorporating geometric or contour-level priors \cite{ravishankar2017learning,wang2022bevt}. Consequently, segmentation outputs may still suffer from boundary leakage, jagged edges, and inconsistent shapes, particularly in challenging echocardiographic scenarios \cite{xie2021segformer,wang2021pyramid}.

Boundary-aware segmentation methods \cite{kervadec2019boundary,qin2019basnet} attempt to improve edge localization by introducing auxiliary boundary supervision or edge-based loss functions. Although these approaches enhance contour sharpness to some extent, boundary information is usually treated as a secondary objective rather than a primary driver of feature refinement. As a result, structural priors are not fully utilized to guide the segmentation process \cite{qin2019basnet}.

In practice, cardiac segmentation is inherently a structural problem \cite{lang2015recommendations}. Clinicians rely on well-defined anatomical contours rather than isolated pixel predictions to assess cardiac function. However, most existing methods predominantly focus on appearance-based learning \cite{litjens2017survey,ronneberger2015u}, with limited integration of explicit structural information. This gap becomes particularly critical under domain shift \cite{guan2021domain}, where appearance features \cite{wang2022generalizing} vary significantly while anatomical structures remain relatively consistent.

To address these limitations, we propose a Contour-Guided Query Refinement Network (CGQR-Net) for boundary-aware cardiac ultrasound segmentation. The key idea is to explicitly incorporate structural priors into the feature refinement process through contour-guided queries. Specifically, a coarse segmentation is first generated, from which anatomical contours are extracted and converted into learnable query embeddings. These contour-derived queries interact with multi-resolution feature representations through cross-attention, enabling structure-aware refinement that improves boundary precision and suppresses noise-induced artifacts. In addition, a dual-head supervision strategy jointly optimizes segmentation and boundary predictions to enforce anatomical consistency.

The proposed framework is evaluated on the Cardiac Acquisitions for Multi-structure Ultrasound Segmentation (CAMUS) dataset \cite{camus} and further validated on the CardiacNet dataset \cite{yangcardiacnet} to assess cross-dataset generalization. The results demonstrate that CGQR-Net achieves improved segmentation accuracy, enhanced boundary delineation, and strong robustness under domain shift, including challenging pathological cases such as atrial septal defect (ASD).

The main contributions of this work are summarized as follows:

\begin{itemize}
\item We introduce a contour-guided query refinement mechanism that transforms segmentation contours into learnable query embeddings, enabling explicit structural guidance during feature refinement.

\item We integrate multi-resolution feature representations with cross-attention to effectively fuse local detail and global context for improved segmentation accuracy.

\item We design a boundary-aware dual-head supervision strategy that jointly optimizes region segmentation and boundary prediction, enhancing anatomical consistency.

\item We demonstrate strong cross-dataset generalization through evaluation on CAMUS and CardiacNet datasets, including both ASD and non-ASD cohorts.
\end{itemize}

% \deleted{The rest of the paper is organised as: Section \ref{relatework} describes a comprehensive review of related work to contextualize our research work followed by proposed method in Section \ref{propose}. In Section \ref{experimental}, we discussed the experimental setup, including datasets and implementation details. Section \ref{results} presents the results, whereas Section \ref{discussion} presents the discussion. Finally, in Section \ref{conclusion}, we present the conclusion.}

The rest of the paper is organized as follows. Section \ref{relatework} reviews the related work. Section \ref{propose} presents the proposed method. Section \ref{experimental} describes the experimental setup, including the datasets and implementation details. Section \ref{results} presents the results. Section \ref{discussion} discusses the findings. Finally, Section \ref{conclusion} concludes the paper.

\section{Related Work}
\label{relatework}

\begin{table*}[t]
\centering
\small
\caption{Summary of existing cardiac segmentation methods, their methodologies, and limitations.}
\label{tab:related_work_summary}
\renewcommand{\arraystretch}{1.2}
\setlength{\tabcolsep}{6pt}

\begin{tabular}{l p{6cm} p{6cm}}
\hline
\textbf{Paper} & \textbf{Methodology} & \textbf{Drawbacks} \\
\hline

U-Net~\cite{ronneberger2015u} 
& Encoder--decoder CNN with skip connections for pixel-wise segmentation. 
& Relies on local features and often produces smooth or inaccurate boundaries in low-contrast ultrasound images. \\

Attention U-Net~\cite{oktay2018attention} 
& Introduces attention gates to focus on relevant regions. 
& Improves region localization but still lacks explicit structural modeling; boundary precision remains limited. \\

UNet++~\cite{zhou2018unet++}
& Dense skip connections for improved multi-scale feature fusion. 
& Increases architectural complexity while remaining primarily driven by pixel-wise learning without explicit contour priors. \\

nnU-Net~\cite{isensee2021nnu} 
& Self-configuring segmentation framework with adaptive design. 
& Provides a strong baseline but remains appearance-driven and struggles with domain shift and boundary ambiguity. \\

CANet~\cite{hanselmann2020canet} 
& Context-aware network leveraging multi-scale contextual information. 
& Produces smoother predictions but lacks precise boundary delineation in challenging regions. \\

Extended nnU-Net~\cite{isensee2023extending} 
& Improved nnU-Net with enhanced configurations and training strategies. 
& Offers improved optimization but still shows limited structural awareness and sensitivity to anatomical variability. \\

DAM-Seg~\cite{ullah2025anatomically} 
& Incorporates anatomical priors using associative memory for improved segmentation. 
& Relies on learned memory representations; contour-level refinement is not explicitly modeled. \\

Tran et al.~\cite{cardiac_fcn} 
& Fully convolutional network trained in an end-to-end manner. 
& Shows weak boundary delineation under fuzzy edge information in ultrasound images. \\

Li et al.~\cite{displacement_aware} 
& Displacement-aware shape encoding to model deformation. 
& May overfit to training domains and shows limited generalization to unseen datasets. \\

Cai et al.~\cite{cross_domain} 
& Cross-domain mixup strategy for domain adaptation. 
& Struggles under large domain gaps and relies more on appearance alignment than structural consistency. \\

Painchaud et al.~\cite{VAE_wrapping} 
& Variational autoencoder for enforcing anatomical plausibility. 
& Requires high memory usage and complex latent representations. \\

Van et al.~\cite{GCN} 
& Combines a CNN encoder with a graph-based decoder for anatomical constraints. 
& Strong structural constraints may fail to capture rare but valid anatomical variations. \\

\textbf{CGQR-Net (Ours)} 
& Multi-resolution feature fusion with contour-guided query refinement and cross-attention. 
& Explicitly integrates contour-derived structural priors for boundary-aware refinement, improving robustness, boundary precision, and cross-dataset generalization. \\

\hline
\end{tabular}
\end{table*}

Table~\ref{tab:related_work_summary} summarizes representative cardiac segmentation methods along with their design strategies and limitations. A clear pattern emerges, for example, most existing approaches are fundamentally appearance-driven, relying on convolutional features or global attention without explicitly modeling anatomical structure. As a result, they commonly suffer from boundary ambiguity, over-smoothing, and degraded robustness under domain shift. Although several works incorporate shape priors or boundary constraints, these are typically imposed as auxiliary supervision rather than integrated into the feature refinement process itself. This exposes a critical gap in the literature: the absence of structure-aware refinement mechanisms that directly leverage contour-level information to guide segmentation.

\subsection{CNN-Based Cardiac Segmentation}

CNN-based encoder-decoder models have dominated cardiac ultrasound segmentation because they learn strong local appearance features and can be trained effectively with limited annotated data \cite{han2026mambaeviscrib,wen2025lpm}. U-Net~\cite{ronneberger2015u} established the standard skip-connected design for dense prediction, while later variants such as Attention U-Net~\cite{oktay2018attention} introduced attention gates to suppress irrelevant activations and improve focus on target regions. In echocardiography, multi-scale context is especially important because the left ventricle, myocardium, and atrium often appear with blurred borders and strong speckle contamination \cite{yan2025multi,aghapanah2025mecardnet}. High-resolution architectures such as HRNet~\cite{wang2020deep} preserve fine spatial information throughout the network and repeatedly fuse representations across scales, which is advantageous for thin and irregular cardiac boundaries. These designs generally improve localization compared with aggressive downsampling backbones \cite{chen2021transunet}.

Despite these advances, CNN-based methods still exhibit a fundamental limitation: they mainly optimize pixel-wise overlap and rely heavily on local convolutional evidence \cite{cui2025p2tc,cui2021multiscale}. Even when multi-scale fusion is used, the models do not explicitly encode contour geometry or anatomical shape priors. Consequently, predictions are often overly smooth, locally fragmented, or anatomically inconsistent, particularly in low-contrast regions where ultrasound artifacts obscure the underlying structures.

\subsection{Transformer-Based Medical Segmentation}

Transformer-based segmentation models were introduced to overcome the locality of convolutions by modeling long-range dependencies through self-attention \cite{li2025aggregate,yan2025multi}. TransUNet~\cite{chen2021transunet} is a representative hybrid design that combines convolutional feature extraction with transformer encoding to improve global context modeling while retaining localization through a U-shaped decoder. Later transformer-based variants further explored hierarchical attention and multi-scale token interaction to improve medical segmentation accuracy. These models are attractive because cardiac structures often require both global shape understanding and local edge precision \cite{chen2021transunet}.

Despite these strengths, most transformer-based methods still operate on image tokens or feature tokens without explicit structural grounding \cite{shamshad2023transformers}. While global attention improves contextual consistency, it does not directly enforce anatomically meaningful boundaries. In echocardiography, where structures may be faint, partially occluded, or corrupted by speckle, unguided token attention remains insufficient for precise boundary recovery \cite{mazher2024self,houssein2024adapting}. In other words, transformer-based models improve global context, but not necessarily boundary refinement guided by structural evidence.

\subsection{Boundary-Aware Segmentation Models}

Boundary-aware learning has been widely explored to improve delineation of thin and ambiguous structures \cite{liu2025fabrf,chen2020contour}. Typical strategies include auxiliary edge branches, boundary attention modules, contour losses, and multi-task learning with explicit edge supervision. In cardiac ultrasound, this is particularly important because small boundary errors can produce non-trivial deviations in derived measurements such as chamber size, wall thickness, and volumetric indices. Recent echocardiography studies have shown that incorporating boundary-related constraints improves segmentation quality, especially when the ventricular wall or atrial boundary is poorly contrasted \cite{zhang2024bsanet,yang2025boundary,zhao2024boundary}.

However, most boundary-aware approaches still treat boundary prediction as an auxiliary task rather than a core component of feature refinement \cite{chen2025dbanet,ji2025bdformer}. Boundary maps are typically predicted in parallel and used only for supervision, which limits their influence on intermediate representations. As a result, segmentation remains largely driven by dense appearance features rather than geometry-aware interactions.

\subsection{Query-Based and Attention-Based Refinement Methods}

Query-based learning has become an important direction in segmentation, especially in transformer-style architectures where learned queries interact with image features through cross-attention \cite{im2025gate3d,tripathi2024query,xu2023multi}. This design enables adaptive grouping of spatial features and iterative refinement of predictions. More recent studies have highlighted a key weakness of standard query-based models: the queries are often randomly initialized or globally learned without domain-specific priors, which can make them unstable or semantically weak in difficult medical images \cite{allahim2025semantic,sugandhika2025vost,cavagnero2024pem}.

This limitation is particularly relevant for ultrasound segmentation \cite{mishra2025tier,ma2024clapsep}. If the query has no geometric grounding, the attention module must infer both anatomy and boundary structure from noisy appearance features alone. That is inefficient and fragile under domain shift. The absence of contour-conditioned queries therefore represents a significant gap in current query-based refinement frameworks.

\subsection{Domain Generalization in Cardiac Imaging}

Domain shift remains a major obstacle in cardiac ultrasound analysis. Differences in scanner vendors, acquisition settings, probe position, frame quality, patient habitus, and pathology can substantially change image appearance and degrade model performance on unseen data. This problem is especially severe in echocardiography, where noise, contrast, and shape variability are much more pronounced than in many other medical imaging modalities. Several studies have explored domain adaptation, self-supervised learning, feature alignment, and multi-task regularization to address this issue \cite{simionescu2025multitask,zacharias2026self,zhang2025multi}. The recently introduced CardiacNet benchmark further emphasizes the need for robust models by reflecting the diversity of echocardiographic abnormalities and acquisition conditions encountered in realistic cardiac data \cite{adiga2024anatomically}.

A fundamental limitation of many domain generalization approaches is their reliance on appearance alignment rather than structural consistency \cite{liu2022single,gu2023cddsa}. While image appearance varies substantially across datasets, anatomical structure changes less dramatically. Methods that ignore this distinction tend to overfit to dataset-specific textures and degrade under distribution shift. This highlights the importance of structural priors as a more robust basis for generalizable cardiac ultrasound segmentation.

\subsection{Summary}

In summary, existing cardiac segmentation methods share a common limitation: they are predominantly appearance-driven and do not explicitly integrate structural priors into feature refinement. CNN-based models rely on local evidence, transformer-based methods improve global context but lack geometric grounding, boundary-aware approaches treat edges as auxiliary signals, and query-based methods typically use generic learnable queries without anatomical conditioning. These limitations collectively lead to suboptimal boundary precision and reduced robustness under domain shift. This reveals a critical gap in the literature: the absence of a unified framework that explicitly integrates contour-derived structural information into the refinement process. Addressing this gap is essential for achieving anatomically consistent and generalizable segmentation in echocardiography.

\begin{figure*}[!t] 
\centering 
\includegraphics[width=1\textwidth]{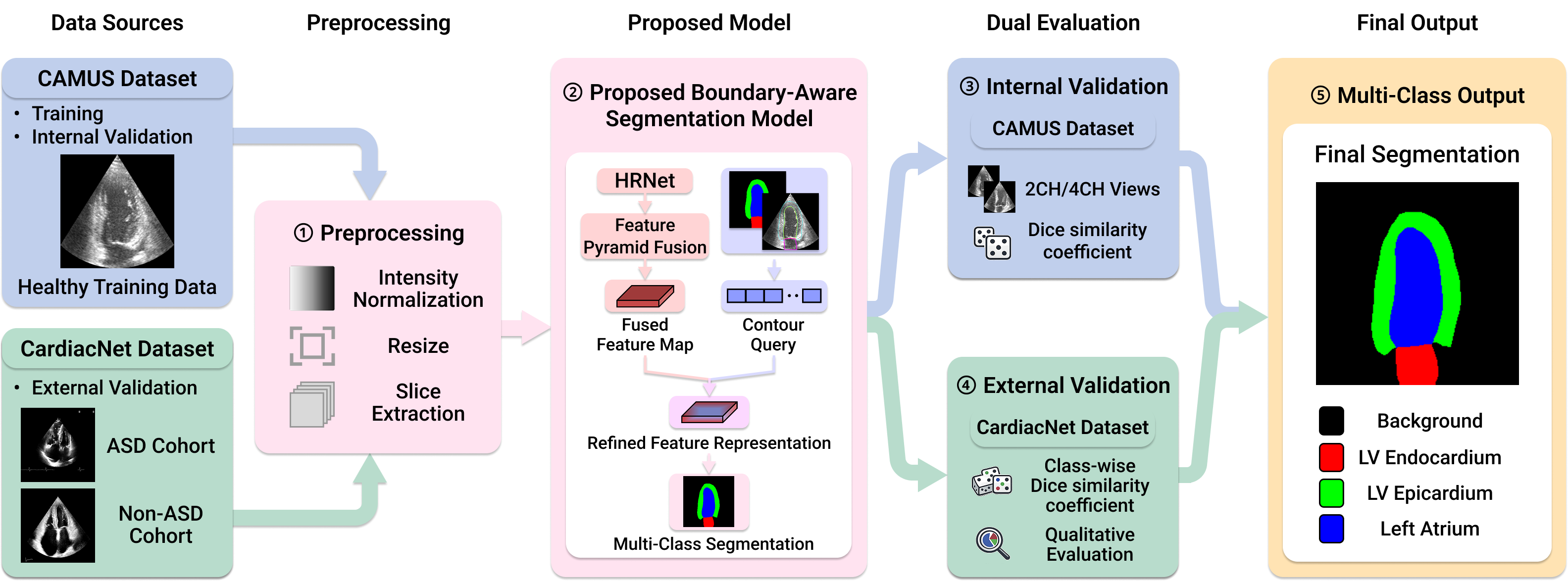} 
\caption{Overall workflow of the proposed framework. CAMUS data, including two-chamber (2CH) and four-chamber (4CH) echocardiographic views, are used for training and internal validation, while CardiacNet is used for external validation. After preprocessing, images are fed into the boundary-aware segmentation model, which integrates HRNet, multi-scale feature fusion, and contour-guided query refinement. The model is evaluated using Dice similarity coefficients and qualitative analysis, producing final multi-class segmentation outputs.} 
\label{workflow} 
\end{figure*}

\section{Proposed Method} 
\label{propose}

\subsection{Overview} 
The proposed CGQR-Net is a boundary-aware multi-class segmentation framework that explicitly injects structural information into feature refinement. As illustrated in Fig.~\ref{workflow}, the framework consists of three stages: preprocessing, coarse structural estimation, and contour-guided refinement. First, input echocardiographic images are normalized and resized to reduce appearance variation across patients and scanners. The processed image is then passed through an HRNet~\cite{wang2020deep} backbone to extract multi-resolution feature maps that preserve both local spatial detail and global anatomical context. HRNet is adopted in this work instead of conventional encoder--decoder backbones because it maintains high-resolution representations throughout the network while repeatedly exchanging information across parallel multi-scale branches. This property is particularly beneficial for echocardiographic segmentation, where weak boundaries, speckle noise, and thin myocardial structures require accurate spatial localization in addition to semantic context. In contrast, architectures that rely on aggressive downsampling may lose fine boundary cues that are critical for contour extraction and subsequent refinement. A coarse segmentation head produces an initial prediction of the cardiac structures. This coarse prediction is not treated as the final output; instead, it serves as a structural hypothesis from which anatomical contours are extracted. These contours are converted into compact query embeddings that summarize boundary geometry. In the proposed method, a \emph{query} is therefore a contour-conditioned structural token rather than a generic learnable vector. It represents geometric information such as contour location, spatial extent, and shape statistics. These query embeddings interact with fused multi-scale image features through cross-attention, allowing the network to selectively enhance boundary-relevant regions and suppress noise-induced ambiguity. Finally, a dual-head prediction module produces both the refined segmentation mask and a boundary map, enabling the model to optimize region overlap and contour consistency simultaneously. 

\subsection{Overall Architecture} 

\begin{figure*}[!ht] 
\centering 
\includegraphics[width=1\textwidth]{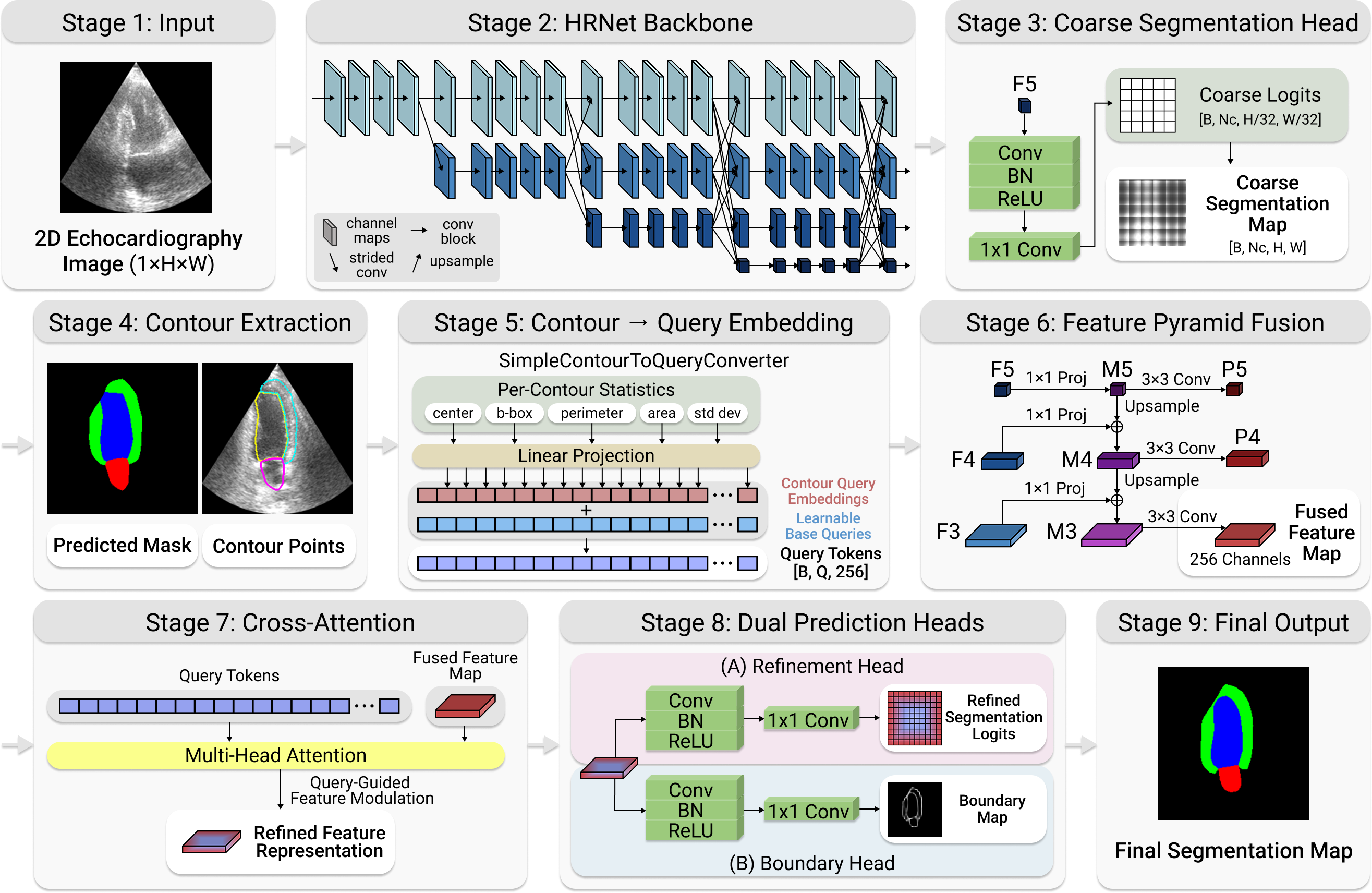} 
\caption{Overview of the proposed contour-guided query refinement framework. A two-dimensional echocardiography image is first processed by a HRNet backbone to extract multi-resolution features, and a coarse segmentation map is then generated. Contour points extracted from the coarse prediction are converted into query embeddings. These contour-derived queries interact with fused multi-scale features through cross-attention, where the queries provide structural guidance and the fused features provide spatial context. The refined representation is finally passed to dual prediction heads to produce the segmentation output and boundary map. Here, 2D denotes two-dimensional, Conv convolution, BN batch normalization, and ReLU rectified linear unit.}
\label{proposed} 
\end{figure*} 

Given an input echocardiographic image $\mathbf{I} \in \mathbb{R}^{1 \times H \times W}$, the proposed CGQR-Net first extracts multi-scale feature representations using an HRNet encoder: 
\begin{equation} 
\{\mathbf{F}_1, \mathbf{F}_2, \mathbf{F}_3\} = \mathcal{E}_{HRNet}(\mathbf{I}),
\label{eq:hrnet_features} 
\end{equation} 
where $\mathcal{E}_{HRNet}(\cdot)$ denotes the multi-resolution feature extractor. These feature maps contain complementary information: shallow branches preserve fine spatial details, while deeper branches encode higher-level semantic context. A coarse segmentation is generated from the deepest feature map: 
\begin{equation} 
\mathbf{Z}_c = \mathbf{W}_c * \mathbf{F}_3 + \mathbf{b}_c, 
\label{eq:coarse_logits} 
\end{equation} 

\begin{equation} 
\mathbf{S}_c = \text{Softmax}(\mathbf{Z}_c), 
\label{eq:coarse_seg} 
\end{equation} 
where $*$ denotes convolution. The coarse prediction $\mathbf{S}_c$ is then used to extract structural contours. These contours are encoded into query embeddings and used to refine the fused multi-scale feature representation through cross-attention. The refined features are passed to two prediction heads to produce the final segmentation map $\mathbf{S}_r$ and the boundary map $\mathbf{B}$. 

\subsection{HRNet Multi-Resolution Feature Encoder} 
Unlike conventional encoder--decoder networks that progressively reduce spatial resolution, HRNet~\cite{wang2020deep} maintains parallel feature streams at multiple resolutions throughout the network. Let $\mathbf{X}^{(k)}$ denote the feature map at the $k$-th resolution branch. The update rule can be written as 

\begin{equation} 
\mathbf{X}^{(k)}_{t+1} = f^{(k)}(\mathbf{X}^{(k)}_{t}) + \sum_{j \neq k} \phi_{j \rightarrow k}(\mathbf{X}^{(j)}_{t}), 
\label{eq:hrnet_update} 
\end{equation} 
where $f^{(k)}(\cdot)$ denotes convolutional transformations within branch $k$, and $\phi_{j \rightarrow k}(\cdot)$ represents cross-scale fusion from branch $j$ to branch $k$. This design is well suited to echocardiographic segmentation because it preserves high-resolution boundary evidence while still incorporating semantic context from lower-resolution branches. 

\subsection{Coarse Segmentation and Contour Extraction} 
The coarse segmentation $\mathbf{S}_c$ provides an initial estimate of each anatomical structure. For class $k$, the predicted foreground region is denoted by $\Omega_k$, and its contour is defined as 

\begin{equation} \mathcal{C}_k = \partial \Omega_k. 
\label{eq:contour_def} 
\end{equation}

Each contour is represented as an ordered set of boundary points: 

\begin{equation} \mathcal{C}_k = \{(x_i, y_i)\}_{i=1}^{N_k},
\label{eq:contour_points} 
\end{equation} 
where $N_k$ is the number of sampled points on the contour. To make the representation invariant to image size, the coordinates are normalized as 

\begin{equation} 
\tilde{x}_i = \frac{x_i}{W}, \qquad \tilde{y}_i = \frac{y_i}{H}. \label{eq:normalize_coords} 
\end{equation} 
This step is important because it converts the dense coarse mask into an explicit structural representation. Instead of relying only on region probabilities, the model now has access to boundary geometry, including location, extent, and contour shape. 

\subsection{Contour-to-Query Embedding Module} 
For each contour $\mathcal{C}_k$, we compute a compact set of geometric descriptors: 

\begin{equation} 
\mu_x = \frac{1}{N_k}\sum_i \tilde{x}_i, \qquad \mu_y = \frac{1}{N_k}\sum_i \tilde{y}_i, 
\label{eq:centroid} 
\end{equation} 

\begin{equation} 
A_k = \text{Area}(\mathcal{C}_k), \qquad P_k = \text{Perimeter}(\mathcal{C}_k), \label{eq:area_perimeter} 
\end{equation}
where $\mu_x$ and $\mu_y$ denote the contour centroid, and $A_k$ and $P_k$ denote its area and perimeter. We further include dispersion terms $(\sigma_x,\sigma_y)$ to encode contour spread. The contour descriptor is therefore 

\begin{equation} \mathbf{d}_k = [\mu_x, \mu_y, A_k, P_k, \sigma_x, \sigma_y]^\top.
\label{eq:descriptor} 
\end{equation} 
The descriptor is projected into the latent feature space to form a query embedding: 
\begin{equation} 
\mathbf{q}_k = \mathbf{W}_q \mathbf{d}_k + \mathbf{b}_q.
\label{eq:query_embedding} 
\end{equation}
In the proposed method, $\mathbf{q}_k$ is the key structural token associated with contour $\mathcal{C}_k$. It is called a \emph{query} because it is used to probe the fused feature map and retrieve boundary-relevant spatial evidence. In addition to contour-derived queries, a small set of learnable base queries $\mathbf{Q}_b \in \mathbb{R}^{Q_b \times d}$ is introduced to provide additional flexibility. The final query matrix is 

\begin{equation} \mathbf{Q} = \{\mathbf{q}_k\}_{k=1}^{K} \cup \mathbf{Q}_b. 
\label{eq:final_queries} 
\end{equation} 

\subsection{Feature Pyramid Fusion} 
The purpose of feature fusion is to combine the complementary strengths of multi-resolution HRNet ~\cite{wang2020deep} branches into a single representation that is rich in both boundary detail and semantic context. Let $\mathbf{F}_1$, $\mathbf{F}_2$, and $\mathbf{F}_3$ denote feature maps from different resolutions. Since these feature maps differ in channel dimension and spatial size, they are first aligned by $1\times1$ convolutions: 

\begin{equation} 
\hat{\mathbf{F}}_i = \psi_i(\mathbf{F}_i), 
\label{eq:channel_align} 
\end{equation} where $\psi_i(\cdot)$ maps each branch to a common channel dimension. Next, all aligned feature maps are upsampled to the same spatial resolution and combined: 
\begin{equation} \mathbf{F}_{fused} = \sum_i \text{Upsample}(\hat{\mathbf{F}}_i). 
\label{eq:fused_features} 
\end{equation} 
This fusion operation is central to the proposed framework. The shallow features contribute fine-grained edge and contour cues, whereas the deeper features provide global anatomical semantics and robustness to local noise. Their fusion yields a unified spatial feature memory that is later attended by the contour-derived queries. In other words, the fused feature map is the image-driven representation, while the queries are the structure-driven representation. 

\subsection{Contour-Guided Cross-Attention Refinement} 
To apply cross-attention, the fused feature map is flattened into a sequence of spatial tokens: 
\begin{equation}
\mathbf{F} \in \mathbb{R}^{N \times d}, \qquad N = H'W', 
\end{equation} 
where $H'$ and $W'$ denote the spatial size of the fused feature map, and $d$ is the embedding dimension. The query matrix is 
\begin{equation} 
\mathbf{Q} \in \mathbb{R}^{M \times d}, 
\end{equation} 
where $M$ is the number of contour-derived and learnable base queries. Cross-attention is performed \emph{between two objects}: (i) the contour-derived query embeddings $\mathbf{Q}$, which provide structural guidance, and (ii) the fused spatial feature tokens $\mathbf{F}$, which provide image evidence and context. The fused features are transformed into keys and values: 
\begin{equation} 
\mathbf{K} = \mathbf{F}\mathbf{W}_K, \qquad \mathbf{V} = \mathbf{F}\mathbf{W}_V, 
\end{equation} 
while the query matrix is projected as 
\begin{equation} 
\mathbf{Q}' = \mathbf{Q}\mathbf{W}_Q. 
\end{equation} 
The attention weights are then computed as 
\begin{equation} \mathbf{A} = \text{Softmax} \left( \frac{\mathbf{Q}'\mathbf{K}^\top}{\sqrt{d}} \right), 
\label{eq:attention_weights} 
\end{equation} 
where $\mathbf{A} \in \mathbb{R}^{M \times N}$ measures how strongly each contour query attends to each spatial location in the fused feature map. The query-conditioned context is obtained as 
\begin{equation} 
\mathbf{H} = \mathbf{A}\mathbf{V}, 
\label{eq:query_context} 
\end{equation}
where $\mathbf{H} \in \mathbb{R}^{M \times d}$. To use this structural information for spatial refinement, the query responses are projected back to the feature domain: 
\begin{equation} 
\mathbf{M} = \mathbf{A}^\top \mathbf{H}\mathbf{W}_M,
\label{eq:modulation_map}
\end{equation}
where $\mathbf{M} \in \mathbb{R}^{N \times d}$ is a query-guided modulation map. The refined feature representation is then computed as 
\begin{equation} 
\mathbf{F}_{ref} = \mathbf{F} + \gamma \mathbf{M}, 
\label{eq:refined_features} 
\end{equation} 
where $\gamma$ is a learnable scaling parameter. This formulation clarifies that the cross-attention is not between two generic token sets; it is specifically between \emph{contour-derived structural queries} and \emph{fused spatial feature tokens}. The resulting feature refinement is therefore explicitly structure-aware. 

\subsection{Boundary-Aware Dual-Head Prediction and Loss} 
The refined feature map is reshaped back to the spatial domain and fed into two prediction heads. The segmentation head produces refined class logits: 
\begin{equation} \mathbf{Z}_r = \mathbf{W}_r * \mathbf{F}_{ref}, 
\label{eq:refined_logits} 
\end{equation} 
and the boundary head predicts edge logits: 
\begin{equation} 
\mathbf{Z}_b = \mathbf{W}_b * \mathbf{F}_{ref}. 
\label{eq:boundary_logits} 
\end{equation} 
The final outputs are 
\begin{equation} \mathbf{S}_r = \text{Softmax}(\mathbf{Z}_r), \qquad \mathbf{B} = \sigma(\mathbf{Z}_b), 
\label{eq:final_outputs} 
\end{equation} 
where $\sigma(\cdot)$ denotes the sigmoid function. The segmentation head focuses on region assignment, while the boundary head emphasizes edge localization. Their joint optimization encourages the final prediction to be both region-consistent and anatomically well delineated. The segmentation loss is defined using multi-class Dice similarity coefficient (DSC) loss: 
\begin{equation} 
\mathcal{L}_{seg} = 1 - \frac{2 \sum_i p_i g_i} {\sum_i p_i + \sum_i g_i + \epsilon}, 
\label{eq:dice_loss} 
\end{equation} 
where $p_i$ and $g_i$ denote predicted and ground-truth pixels, respectively. Boundary supervision is imposed through binary cross-entropy: 
\begin{equation} 
\mathcal{L}_{boundary} = -\sum_i \left[ b_i \log \hat{b}_i + (1-b_i)\log(1-\hat{b}_i) \right]. 
\label{eq:bce_loss}
\end{equation} 
The total objective is 
\begin{equation} \mathcal{L} = \mathcal{L}_{seg} + \lambda \mathcal{L}_{boundary}, 
\label{eq:total_loss} 
\end{equation}
where $\lambda$ balances region-level and boundary-level supervision.

%%%%%%%%%%%%%%%%%%%%%%%

\section{Experimental Setup}
\label{experimental}

\subsection{Datasets}

\subsubsection{CAMUS Dataset}

The CAMUS dataset \cite{camus} is a publicly available benchmark for echocardiographic segmentation. It contains annotated two-dimensional transthoracic echocardiography images acquired in both two-chamber (2CH) and four-chamber (4CH) views. For each subject, end-diastolic (ED) and end-systolic (ES) frames are provided, corresponding to the maximum and minimum ventricular volumes within the cardiac cycle.

Manual annotations include three primary anatomical structures: LV endocardium, LV epicardium, and LA. Let $\Omega_k$ denote the region corresponding to class $k$, where $k \in \{1,2,3\}$ represents LV endocardium, LV epicardium, and LA, respectively, and $k=0$ denotes the background class.

For each image $\mathbf{I}$, the corresponding ground-truth mask $\mathbf{G} \in \{0,1,2,3\}^{H \times W}$ encodes pixel-wise class labels. Both ED and ES phases are evaluated independently to assess segmentation performance across cardiac motion variability. The inclusion of both 2CH and 4CH views ensures robustness to anatomical appearance variations across acquisition planes.

\subsubsection{CardiacNet Dataset}

To evaluate cross-dataset generalization and robustness under pathological variation, we further validate the proposed method on the CardiacNet dataset \cite{yangcardiacnet}. This dataset contains echocardiographic images from two distinct cohorts: atrial septal defect (ASD) patients and non-ASD subjects. The ASD cohort exhibits structural abnormalities that introduce additional morphological variability, making segmentation more challenging. CardiacNet serves as an external validation dataset. No joint training or domain adaptation strategy is applied between CAMUS and CardiacNet, allowing direct assessment of domain generalization capability. Let $\mathcal{D}_{C}$ denote the CAMUS dataset and $\mathcal{D}_{N}$ denote the CardiacNet dataset. The model is trained on $\mathcal{D}_{C}$ and evaluated on $\mathcal{D}_{N}$ to quantify performance under distribution shift. This evaluation protocol enables analysis of the structural robustness of contour-guided refinement across heterogeneous acquisition conditions and pathological cases. Fig. \ref{datasets_sample} illustrates representative sample images for CAMUS and CardiacNet datasets.

\subsection{Preprocessing}

Prior to training, all images undergo standardized preprocessing to reduce intensity variation and ensure consistent spatial resolution.

\subsubsection{Intensity Normalization}

Given an input image $\mathbf{I}$, intensity normalization is performed using per-image standardization:
\begin{equation}
\tilde{\mathbf{I}} =
\frac{\mathbf{I} - \mu_{\mathbf{I}}}
{\sigma_{\mathbf{I}} + \epsilon},
\end{equation}
where $\mu_{\mathbf{I}}$ and $\sigma_{\mathbf{I}}$ denote the mean and standard deviation of pixel intensities, and $\epsilon$ is a small constant to prevent numerical instability. This normalization mitigates variability caused by different ultrasound machines and acquisition settings.

\subsubsection{Spatial Resizing}

All images and masks are resized to a fixed spatial resolution $(H_0, W_0)$ using bilinear interpolation for images and nearest-neighbor interpolation for segmentation masks:
\begin{equation}
\mathbf{I}_{res} = \mathcal{R}(\tilde{\mathbf{I}}),
\quad
\mathbf{G}_{res} = \mathcal{R}_{nn}(\mathbf{G}),
\end{equation}
where $\mathcal{R}(\cdot)$ and $\mathcal{R}_{nn}(\cdot)$ denote resizing operators. This ensures uniform tensor dimensions for batch processing and stable multi-scale feature extraction.

\subsubsection{Slice Extraction}

For volumetric sequences or multi-frame acquisitions, individual 2D frames are extracted and treated as independent training samples. Let a sequence be denoted as $\{\mathbf{I}_t\}_{t=1}^{T}$. Each frame is processed independently:
\begin{equation}
\mathcal{S} = \{\mathbf{I}_t, \mathbf{G}_t\}_{t=1}^{T}.
\end{equation}
Frames containing no foreground pixels are optionally excluded to avoid bias toward background-dominant samples.

\subsubsection{Train and Validation Split}

For the CAMUS dataset, training and validation sets are constructed following the official protocol or an 80\%-20\% split at the patient level to avoid data leakage. Let $\mathcal{D}_{train}$ and $\mathcal{D}_{val}$ denote the training and validation sets, respectively:
\begin{equation}
\mathcal{D}_{C} =
\mathcal{D}_{train}
\cup
\mathcal{D}_{val},
\quad
\mathcal{D}_{train}
\cap
\mathcal{D}_{val}
=
\emptyset.
\end{equation}

For CardiacNet, the dataset is partitioned similarly into internal training and validation subsets when training directly on CardiacNet, or used solely as an external test set when evaluating cross-dataset generalization.

This preprocessing pipeline ensures consistency across datasets and enables fair evaluation of the proposed contour-guided segmentation framework.

\subsection{Implementation Details}

All experiments were implemented in Python using the PyTorch deep learning framework. The proposed network was trained using the AdamW optimizer, which combines adaptive gradient estimation with decoupled weight decay regularization. The initial learning rate was set to $1 \times 10^{-4}$.

A cosine annealing learning rate scheduler was applied to progressively reduce the learning rate over training epochs. The batch size was set to 4, and the network was trained for 100 epochs for each cardiac phase (ED and ES). During training, teacher-forcing refinement was applied for the first 20 epochs to stabilize contour-guided query learning.

The number of segmentation classes was set to four, including background, LV endocardium, LV epicardium, and left atrium. Training and inference were performed using an NVIDIA GPU with CUDA acceleration. All experiments were conducted on a workstation equipped with a high-performance RTX 3090 GPU, and 64 GB of RAM. Model parameters were initialized using PyTorch default initialization. The best-performing model was selected based on validation DSC and saved for evaluation. The hyperparameters used in the proposed model are illustrated in Table \ref{tab:hyperparameters}. 

\begin{table}[!t]
\caption{Training hyperparameters used for model training.}
\centering
\normalsize
\begin{tabular}{lc}
\hline
\textbf{Parameter} & \textbf{Value} \\
\hline
Framework & PyTorch \\
Optimizer & AdamW \\
Initial Learning Rate & $1 \times 10^{-4}$ \\
Learning Rate Scheduler & Cosine Annealing \\
Batch Size & 4 \\
Number of Epochs & 100 \\
Teacher-Forcing Epochs & 20 \\
Number of Classes & 4 \\
Loss Function & DSC + Boundary Loss \\
Hardware & NVIDIA GPU (CUDA) \\
\hline
\end{tabular}
\label{tab:hyperparameters}
\end{table}

\subsection{Evaluation Metrics}

Segmentation performance was evaluated using DSC, which measures the overlap between predicted segmentation and ground truth. For a predicted mask $\mathbf{P}$ and ground truth mask $\mathbf{G}$, DSC is defined as

\begin{equation}
DSC =
\frac{2|\mathbf{P} \cap \mathbf{G}|}
{|\mathbf{P}| + |\mathbf{G}|}.
\end{equation}

For multi-class segmentation, DSC is computed independently for each class $k$:

\begin{equation}
\mathrm{DSC}_k, =
\frac{2 \sum_i p_{k,i} g_{k,i}}
{\sum_i p_{k,i} + \sum_i g_{k,i} + \epsilon},
\end{equation}

where $p_{k,i}$ and $g_{k,i}$ denote predicted and ground-truth pixels for class $k$ at location $i$, and $\epsilon$ is a small constant for numerical stability.

The overall DSC is computed as the mean across all foreground classes:

\begin{equation}
\mathrm{DSC}_{\text{mean}}
=
\frac{1}{K}
\sum_{k=1}^{K}
\mathrm{DSC}_k,
\end{equation}

where $K$ denotes the number of anatomical classes.

The training process is illustrated in algorithm \ref{alg:training}.

\begin{algorithm}[t]
\caption{Phase-Specific Training of CGQR-Net}
\label{alg:training}
\footnotesize
\begin{algorithmic}[1]

\State \textbf{Input:} Dataset $\mathcal{D}$, epochs $E$, teacher-forcing epochs $E_{TF}$, batch size $B$, learning rate $\eta$
\State \textbf{Output:} Best model weights per phase

\For{each cardiac phase $p \in \{\text{ED}, \text{ES}\}$}

    \State Split $\mathcal{D}$ into training $\mathcal{D}_{train}^p$ and validation $\mathcal{D}_{val}^p$
    \State Initialize model $\mathcal{M}$
    \State Initialize AdamW optimizer with learning rate $\eta$
    \State Initialize cosine annealing scheduler
    \State $\mathrm{DSC}_{\text{best}} \gets 0$
    \For{$e = 1$ to $E$}

        \For{each mini-batch $(\mathbf{I}, \mathbf{G})$ in $\mathcal{D}_{train}^p$}

            \If{$e \leq E_{TF}$}
                \State Forward pass with teacher forcing
            \Else
                \State Standard forward pass
            \EndIf

            \State Compute refined logits $\mathbf{Z}_r$
            \State Compute boundary logits $\mathbf{Z}_b$

            \State Compute total loss:
            \[
            \mathcal{L} = \mathcal{L}_{seg} + \lambda \mathcal{L}_{boundary}
            \]

            \State Backpropagate gradients
            \State Update model parameters

        \EndFor

        \State Validate model and compute DSC
        \State Update learning rate scheduler

        \If{$\mathrm{DSC}_{\mathrm{val}} > \mathrm{DSC}_{\mathrm{best}}$}
            \State Save model weights
            \State $\mathrm{DSC}_{\mathrm{best}} \gets \mathrm{DSC}_{\mathrm{val}}$
        \EndIf

    \EndFor

\EndFor

\end{algorithmic}
\end{algorithm}

\section{Results}
\label{results}

\subsection{Quantitative Results on CAMUS}
Table \ref{tab:camus_comparison} presents the quantitative comparison of the proposed CGQR-Net with several state-of-the-art segmentation methods on the CAMUS dataset in terms of DSC (\%).

The proposed method achieves the best overall performance with an average DSC of 90.40\%, outperforming all competing approaches. CGQR-Net consistently improves segmentation accuracy across all cardiac structures, including the endocardium, epicardium, and left atrium, demonstrating its robustness in handling different anatomical regions.

For the endocardium, CGQR-Net achieves a DSC of 93.95\%, slightly surpassing strong baselines such as CANet (93.91\%) and DAM-Seg (93.72\%). Although the margin is relatively small, this indicates that the proposed model maintains high accuracy even for structures that are comparatively easier to segment. In contrast, more noticeable improvements are observed for challenging regions. For the epicardium, which suffers from low contrast and weak boundary definition, CGQR-Net attains a DSC of 88.09\%, outperforming DAM-Seg (88.00\%) and CANet (87.52\%). This improvement highlights the ability of the proposed method to better capture ambiguous myocardial boundaries.

The most significant gain is achieved in the segmentation of the left atrium, where CGQR-Net obtains a DSC of 89.15\%, exceeding DAM-Seg (89.10\%) and showing a clear improvement over CANet (87.01\%). The left atrium is particularly difficult to segment due to its irregular shape and high variability across patients. The superior performance of CGQR-Net in this region demonstrates its effectiveness in modeling complex anatomical structures.

Compared with classical architectures such as U-Net, Attention U-Net, and UNet++, the proposed method shows consistent improvements across all structures, especially in boundary-sensitive regions. Furthermore, CGQR-Net outperforms nnU-Net and its extended variant, which are widely regarded as strong baselines in medical image segmentation. These improvements can be attributed to the integration of an HRNet backbone that preserves detailed spatial information, a contour-guided query embedding mechanism that encodes structural priors, and a cross-attention refinement strategy that enhances feature representation through effective interaction between queries and multi-scale features.

\begin{table}[t]
\centering
\caption{Comparison of cardiac structure segmentation performance on the CAMUS dataset in terms of DSC (\%). The best results are shown in bold.}
\label{tab:camus_comparison}
\renewcommand{\arraystretch}{1.15}
\setlength{\tabcolsep}{6pt}
\resizebox{\columnwidth}{!}{%
\begin{tabular}{lcccc}
\hline
\textbf{Model} & \textbf{Endocardium} & \textbf{Epicardium} & \textbf{Left Atrium} & \textbf{Avg.} \\
\hline

U-Net \cite{ronneberger2015u}               & 93.69 & 86.25 & 85.34 & 88.43 \\
Attention U-Net \cite{oktay2018attention}      & 93.72 & 86.54 & 86.02 & 88.76 \\
UNet++ \cite{zhou2018unet++}               & 93.81 & 86.88 & 86.47 & 89.05 \\
nnU-Net \cite{isensee2021nnu}              & 93.33 & 87.06 & 85.13 & 88.51 \\
CANet \cite{hanselmann2020canet}                & 93.91 & 87.52 & 87.01 & 89.48 \\
Extended nnU-Net \cite{isensee2023extending}     & 92.90 & 85.81 & 86.54 & 88.42 \\
DAM-Seg \cite{ullah2025anatomically}               & 93.72 & 88.00 & 89.10 & 90.27 \\

\textbf{CGQR-Net (Ours)} 
                      & \textbf{93.95} 
                      & \textbf{88.09} 
                      & \textbf{89.15} 
                      & \textbf{90.40} \\

\hline
\end{tabular}}
\end{table}

Overall, the results demonstrate that CGQR-Net not only improves segmentation accuracy but also provides better boundary delineation and structural consistency, making it highly suitable for echocardiographic image segmentation tasks.

\subsection{External Validation on CardiacNet}
To evaluate the generalization capability of the proposed method, we conduct external validation on the CardiacNet dataset. Table \ref{tab:cardiacnet_comparison} presents the quantitative comparison of CGQR-Net with several state-of-the-art segmentation methods in terms of DSC (\%).

\begin{table*}[t]
\centering
\caption{Comparison of segmentation performance on the CardiacNet dataset in terms of DSC (\%). The best results are shown in bold.}
\label{tab:cardiacnet_comparison}
\renewcommand{\arraystretch}{1.15}
\setlength{\tabcolsep}{6pt}

\begin{tabular}{lccccc}
\hline
\textbf{Model} & \textbf{Left Atrium} & \textbf{Right Atrium} & \textbf{Left Ventricle} & \textbf{Right Ventricle} & \textbf{Avg.} \\
\hline

U-Net \cite{ronneberger2015u}           & 88.61 & 87.98 & 91.51 & 88.19 & 89.07 \\
Attention U-Net \cite{oktay2018attention}  & 88.94 & 88.31 & 91.63 & 88.54 & 89.36 \\
UNet++ \cite{zhou2018unet++}          & 89.05 & 88.42 & 91.68 & 88.71 & 89.47 \\
nnU-Net \cite{isensee2021nnu}         & 88.66 & 88.47 & 91.40 & 88.50 & 89.26 \\
CANet  \cite{hanselmann2020canet}          & 88.95 & 88.91 & 91.70 & 89.26 & 89.71 \\
Extended nnU-Net \cite{isensee2023extending} & 88.21 & 88.23 & 91.03 & 88.28 & 88.94 \\
DAM-Seg \cite{ullah2025anatomically}         & 89.52 & 87.68 & 91.84 & 89.82 & 89.71 \\

\textbf{CGQR-Net (Ours)}
                 & \textbf{89.60}
                 & \textbf{89.73}
                 & \textbf{91.87}
                 & \textbf{89.89}
                 & \textbf{90.27} \\

\hline
\end{tabular}
\end{table*}

The proposed CGQR-Net achieves the best overall performance with an average DSC of 90.27\%, outperforming all competing methods across all cardiac structures. Notably, the model demonstrates consistent improvements in both atrial and ventricular segmentation tasks. For the left atrium, CGQR-Net attains a DSC of 89.60\%, surpassing strong baselines such as DAM-Seg (89.52\%) and CANet (88.95\%). Similarly, for the right atrium, the proposed method achieves 89.73\%, which is significantly higher than all compared methods, indicating improved robustness in handling complex atrial structures.

For the left ventricle, CGQR-Net achieves 91.87\%, which is the highest among all methods, slightly improving upon DAM-Seg (91.84\%) and outperforming nnU-Net (91.40\%). In the case of the right ventricle, which is typically more challenging due to shape variability and boundary ambiguity, CGQR-Net achieves 89.89\%, exceeding DAM-Seg (89.82\%) and CANet (89.26\%). These consistent improvements across all four structures demonstrate the effectiveness of the proposed contour-guided refinement mechanism in capturing both global structure and fine-grained boundary details.

From a domain generalization perspective, the performance gains on CardiacNet are particularly significant. Unlike CAMUS, the CardiacNet dataset exhibits greater variability in terms of imaging conditions, noise distribution, and anatomical appearance, which introduces a noticeable domain shift. Traditional CNN-based models such as U-Net and UNet++ show degraded performance under this shift, highlighting their limited ability to generalize beyond the training domain. Although nnU-Net provides a strong baseline due to its adaptive design, its performance is still inferior to the proposed method.

The superior generalization ability of CGQR-Net can be attributed to its contour-guided query refinement strategy, which explicitly encodes structural priors derived from coarse segmentation. By transforming contour information into query embeddings and leveraging cross-attention for feature refinement, the model becomes less sensitive to intensity variations and noise patterns across datasets. In addition, the HRNet backbone preserves high-resolution spatial information, enabling the model to maintain consistent performance even under significant domain discrepancies.

Overall, the results on the CardiacNet dataset demonstrate that CGQR-Net not only achieves state-of-the-art segmentation accuracy but also exhibits strong domain generalization capability. This makes the proposed method particularly suitable for real-world clinical deployment, where variability across imaging devices and patient populations is inevitable.

\subsection{Qualitative Results}
\begin{figure*}[!ht]
     \centering
     \includegraphics[width=1\textwidth]{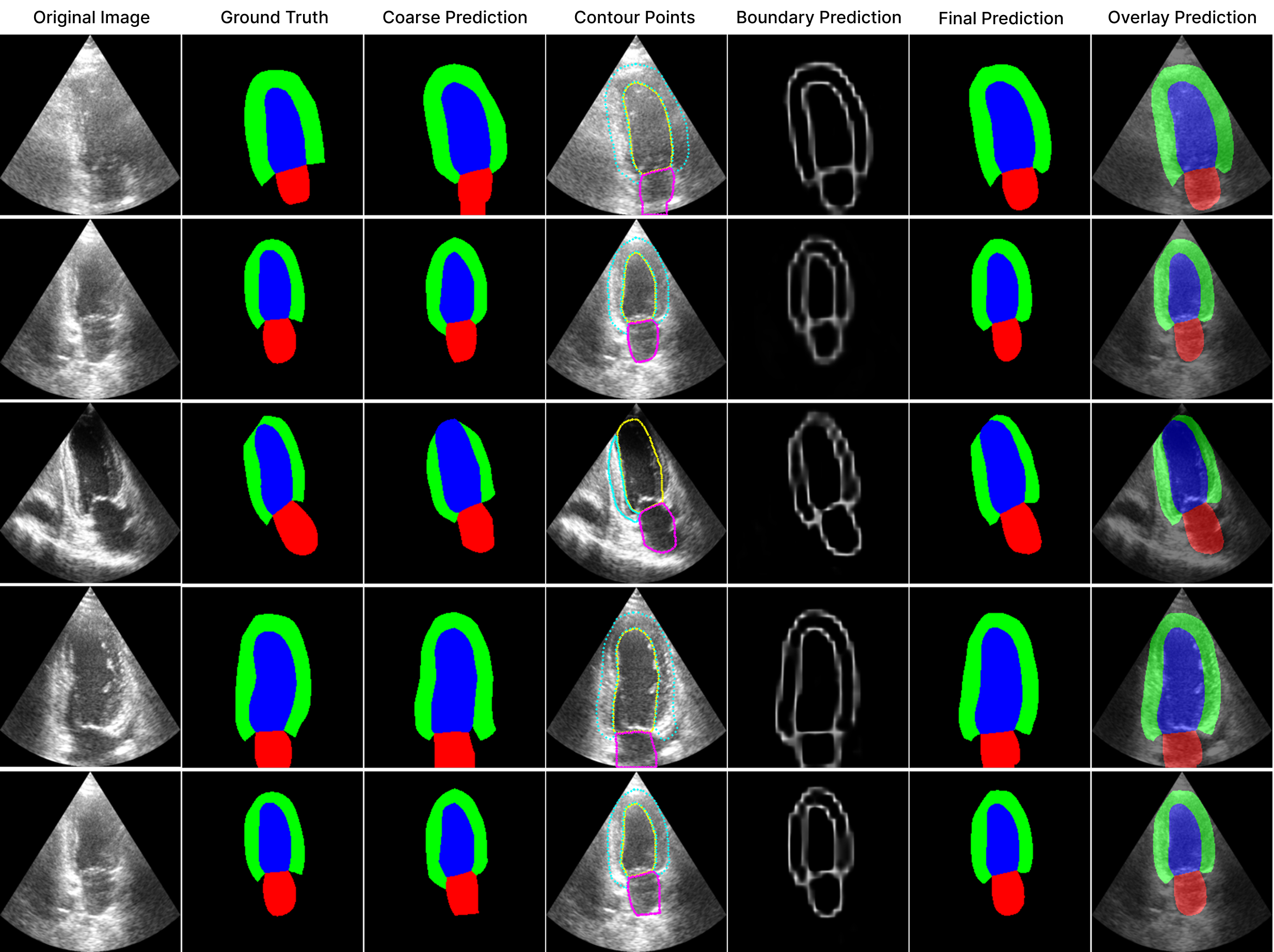}
     \caption{Qualitative segmentation results on the CAMUS dataset. From left to right: original image, ground truth, coarse prediction, contour points, boundary map, final prediction, and overlay visualization. The proposed method produces more accurate and sharper boundaries compared to the coarse segmentation.}
     \label{camus_visual}
 \end{figure*}

\begin{figure*}[!ht]
     \centering
     \includegraphics[width=1\textwidth]{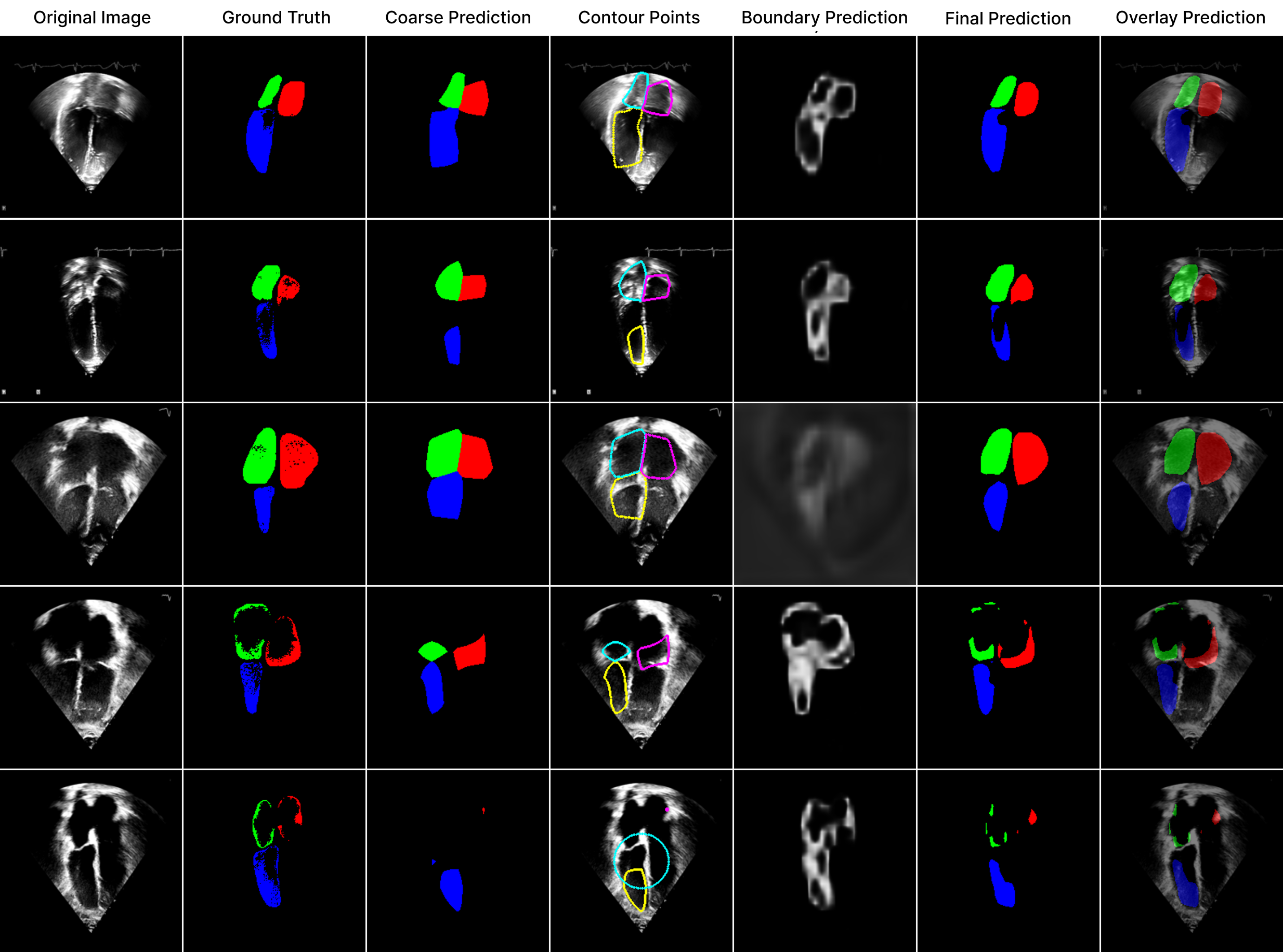}
     \caption{Qualitative segmentation results on the CardiacNet dataset. From left to right: original image, ground truth, coarse prediction, contour points, boundary map, final prediction, and overlay visualization. The proposed method shows improved robustness under challenging conditions, producing more accurate and consistent segmentations with better boundary delineation.}
     \label{cardiac_visual}
 \end{figure*}

Fig. \ref{camus_visual} and Fig. \ref{cardiac_visual} present qualitative segmentation results on the CAMUS and CardiacNet datasets, respectively, illustrating the effectiveness of the proposed CGQR-Net in comparison with ground truth annotations. Each example includes the original image, coarse prediction, contour points, boundary map, final prediction, and overlay visualization.

From the visual results on the CAMUS dataset, it can be observed that the coarse segmentation provides a reasonable initialization but often suffers from inaccurate boundaries and slight structural distortions, particularly around the epicardium and left atrium regions. The extracted contour points capture essential structural priors, highlighting key anatomical boundaries. These contour-guided queries enable the model to refine ambiguous regions effectively. As a result, the final predictions exhibit improved alignment with the ground truth, especially along thin and irregular boundaries. The boundary maps further demonstrate that the model successfully learns sharp and continuous edge representations, which directly contribute to the improved segmentation quality.

On the CardiacNet dataset, which exhibits more severe noise, intensity variations, and anatomical complexity, the advantages of the proposed method become more pronounced. The coarse predictions in these cases often show fragmented regions and boundary leakage due to domain shift. However, after applying contour-guided query refinement, the final segmentation results become significantly more coherent and anatomically consistent. The boundary predictions clearly highlight object edges even under challenging imaging conditions, indicating that the model effectively captures structural information despite noise and low contrast.

Compared to conventional segmentation behavior, where predictions tend to be overly smooth or misaligned at object boundaries, CGQR-Net produces sharper and more accurate delineations. This is particularly evident in difficult cases with weak contrast or irregular shapes, where the proposed method reduces boundary ambiguity and improves shape preservation. The overlay visualizations further confirm that the refined predictions closely match the underlying anatomical structures.

Overall, the qualitative results demonstrate that the proposed contour-guided query mechanism plays a critical role in enhancing boundary localization and structural consistency. The integration of contour information with cross-attention refinement allows the model to correct coarse errors and produce more precise segmentation outputs, validating the effectiveness of the proposed approach in both standard and cross-domain scenarios.

\subsection{Comparison with Existing Methods}
\begin{figure*}[!ht]
     \centering
     \includegraphics[width=1\textwidth]{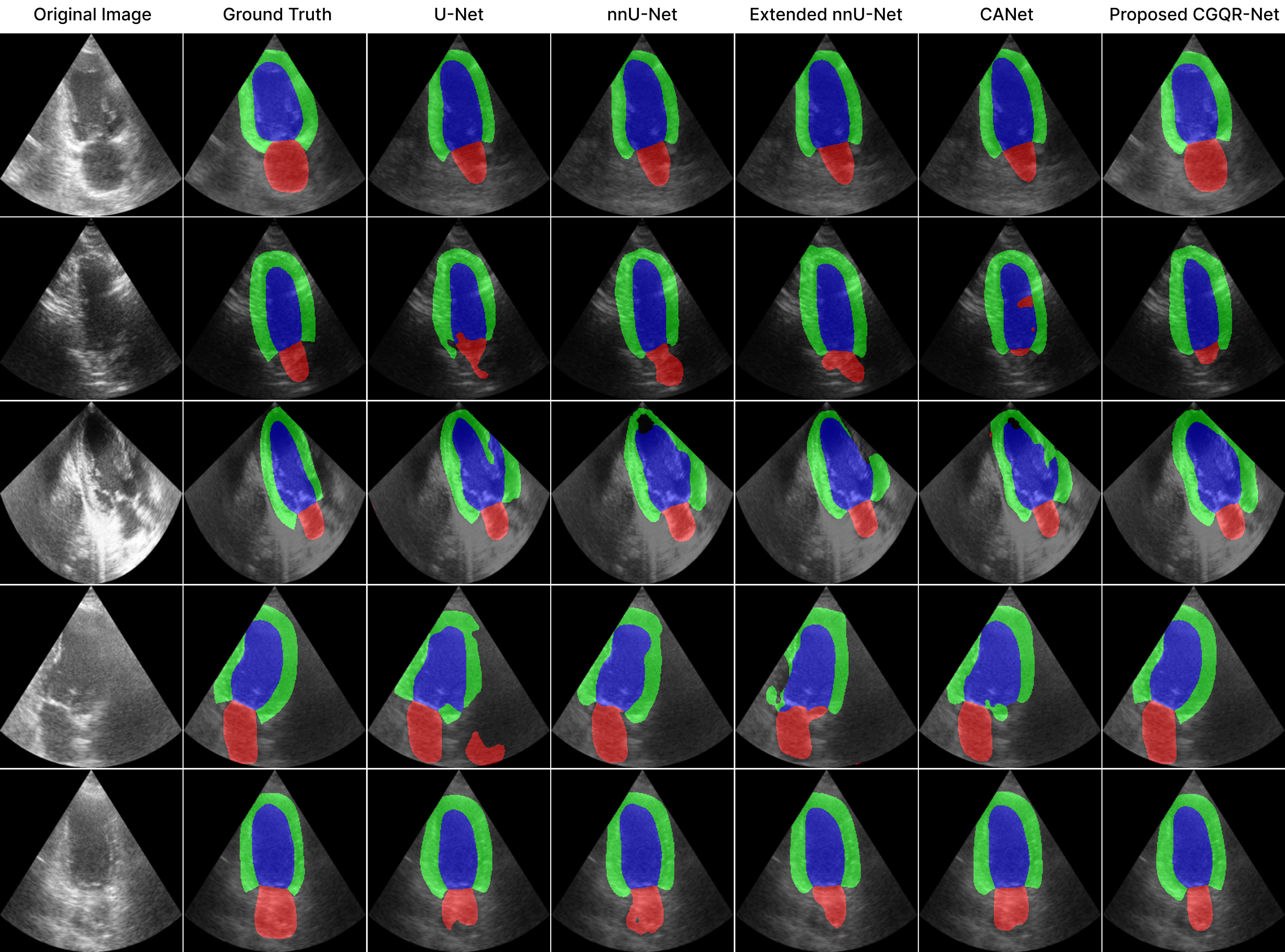}
     \caption{Visual comparison of segmentation results with existing methods on the CAMUS dataset. From left to right: original image and predictions from different methods, including the proposed CGQR-Net. The proposed method produces more accurate segmentation with improved boundary delineation and better shape consistency compared to previous approaches.}
     \label{comparison_camus}
     \end{figure*}

Fig. \ref{comparison_camus} presents a visual comparison between the proposed CGQR-Net and existing segmentation methods on representative echocardiographic samples. The comparison clearly highlights the limitations of previous approaches and demonstrates the effectiveness of the proposed contour-guided refinement strategy.

Conventional CNN-based methods such as U-Net and its variants exhibit reasonably good performance for the endocardium but struggle significantly with boundary precision. As observed in the visual results, these models tend to produce overly smooth segmentations and often fail to accurately delineate thin myocardial regions. In several cases, the epicardium is either under-segmented or merged with surrounding structures, indicating weak boundary awareness. Additionally, leakage artifacts are visible near the left atrium, particularly in low-contrast regions, where the model fails to distinguish between adjacent anatomical structures.

Attention-based models show slight improvements in focusing on relevant regions; however, they still suffer from inconsistent predictions. The segmentation outputs frequently contain fragmented regions or disconnected components, especially in challenging cases with strong speckle noise. These models lack explicit structural guidance, which results in unstable boundary predictions and poor shape consistency.

Methods such as nnU-Net and its extended variants provide more robust performance due to adaptive configurations, but their predictions remain limited by purely pixel-wise learning. As seen in the figure, these methods often produce coarse boundaries and fail to capture fine anatomical details. In particular, irregular structures such as the left atrium exhibit noticeable shape distortion, and boundary transitions appear blurred, indicating insufficient modeling of structural priors.

More advanced approaches like CANet and DAM-Seg improve segmentation quality by incorporating contextual or memory-based mechanisms. While these methods produce smoother predictions, they still struggle with precise boundary localization. The visual results reveal that these models tend to over-smooth the segmentation, leading to loss of fine details and inaccurate delineation in regions with complex geometry. In difficult cases, partial segmentation or missing regions can still be observed, especially under noisy imaging conditions.

In contrast, the proposed CGQR-Net produces significantly more accurate and consistent segmentation results across all samples. The integration of contour-guided query refinement enables the model to explicitly utilize structural information derived from coarse predictions. As a result, the refined outputs exhibit sharper boundaries, better alignment with anatomical structures, and improved shape preservation. Unlike previous methods, CGQR-Net effectively reduces leakage artifacts and avoids over-smoothing, particularly in the epicardium and left atrium regions.

Furthermore, the proposed method demonstrates superior robustness in challenging scenarios, including low contrast, speckle noise, and irregular anatomical shapes. The boundary predictions clearly highlight well-defined edges, which directly contribute to improved segmentation accuracy. This indicates that CGQR-Net not only enhances pixel-wise classification but also enforces structural consistency through contour-based guidance.

Overall, the visual comparison confirms that existing methods are limited by their inability to explicitly model boundary and shape information. In contrast, CGQR-Net successfully addresses these limitations by combining contour extraction, query-based embedding, and cross-attention refinement, resulting in more precise and reliable segmentation outcomes.

\subsection{Ablation Studies}

To evaluate the contribution of each component in the proposed CGQR-Net, we conducted a comprehensive ablation study on the CAMUS dataset. Starting from the full model, we progressively removed or modified key modules, including the boundary supervision branch, coarse segmentation head, contour-guided query mechanism, multi-scale feature pyramid fusion, and teacher-forcing strategy. Table~\ref{tab:ablation_classwise} summarizes the class-wise DSCs for each configuration.

\subsubsection{Effect of Boundary-Aware Supervision}

Removing the boundary head results in a noticeable performance drop across all classes. In particular, LV epicardium DSC decreases from 88.09\% to 87.34\%, and LA DSC drops from 89.15\% to 88.42\%. This confirms that explicit boundary modeling enhances structural consistency, especially for thin and irregular myocardial borders.

\subsubsection{Effect of Coarse Segmentation Head}

When the coarse segmentation head is removed, overall performance slightly degrades (e.g., LV epicardium DSC reduces to 87.76\%). The coarse prediction provides global structural guidance that stabilizes contour extraction and query initialization, improving refinement robustness.

\subsubsection{Effect of Contour-Guided Query Mechanism}

Eliminating the contour-query module produces the most significant degradation. LV endocardium DSC decreases from 93.95\% to 92.87\%, while LV epicardium DSC drops from 88.09\% to 86.95\%. This demonstrates that contour-derived structural priors play a critical role in guiding cross-attention refinement and enhancing boundary precision.

\subsubsection{Effect of Multi-Scale Feature Fusion}

Removing multi-resolution feature pyramid fusion leads to consistent reductions across all classes (LV endocardium: 92.65\%, LV epicardium: 86.71\%, LA: 87.54\%). This confirms that multi-scale aggregation is essential for preserving both global context and fine-grained anatomical details.

\subsubsection{Effect of Teacher-Forcing Warm-Up}

Disabling teacher forcing during early training results in moderate degradation (LV endocardium: 93.02\%, LV epicardium: 87.11\%). Teacher-forcing stabilization improves convergence of contour-query learning during initial epochs, leading to better refinement consistency.

Among all components, the contour-guided query refinement mechanism contributes the largest performance gain, validating the core design principle of integrating geometric structural priors into attention-based refinement. Boundary-aware supervision and multi-resolution fusion further enhance anatomical consistency and segmentation robustness.

These results collectively confirm that each module in CGQR-Net contributes positively to final segmentation accuracy, with the full model achieving the best class-wise DSCs of 93.95\% (LV endocardium), 88.09\% (LV epicardium), and 89.15\% (LA).

\begin{table}[t]
\caption{Class-wise ablation study of CGQR-Net on the CAMUS dataset (\% DSC)}
\centering
\small
\renewcommand{\arraystretch}{1.1}
{
\begin{tabular*}{\columnwidth}{@{\extracolsep{\fill}}lccc@{}}
\hline
\textbf{Method} & \textbf{LV Endo.} & \textbf{LV Epi.} & \textbf{LA} \\
\hline
Full CGQR-Net              & \textbf{93.95} & \textbf{88.09} & \textbf{89.15} \\
w/o Boundary Head          & 93.21 & 87.34 & 88.42 \\
w/o Coarse Head            & 93.48 & 87.76 & 88.63 \\
w/o Contour Queries        & 92.87 & 86.95 & 87.88 \\
w/o Feature Pyramid Fusion & 92.65 & 86.71 & 87.54 \\
w/o Teacher Forcing        & 93.02 & 87.11 & 88.06 \\
\hline
\end{tabular*}
}

\vspace{2pt}
\parbox{\columnwidth}{\footnotesize\textit{Note:} Endo., endocardium; Epi., epicardium.}
\label{tab:ablation_classwise}
\end{table}

\subsection{Cross-Dataset Generalization Analysis}
The experimental results demonstrate that the proposed CGQR-Net exhibits strong generalization capability across datasets with significantly different characteristics. In particular, the performance consistency between the CAMUS and CardiacNet datasets indicates that the model is less sensitive to domain shift compared to conventional approaches. This robustness can be primarily attributed to the contour-guided query refinement mechanism, which enables the model to rely more on structural information rather than purely intensity-based features.

One of the key advantages of contour guidance is its ability to encode anatomical structure explicitly. Unlike pixel-wise learning, which is highly dependent on image appearance, contour-based representations capture the geometric shape and boundary information of cardiac structures. Since anatomical shapes remain relatively consistent across datasets, even when imaging conditions vary, the use of contour-derived queries allows the model to focus on invariant structural features. This significantly improves robustness in the presence of variations such as noise, contrast differences, and imaging artifacts.

Furthermore, the proposed method reduces sensitivity to domain shift by decoupling structural learning from intensity distributions. Traditional CNN-based models tend to overfit to dataset-specific intensity patterns, which leads to performance degradation when applied to unseen data. In contrast, CGQR-Net leverages contour-guided queries that are extracted from coarse segmentation and refined through cross-attention. This process emphasizes spatial relationships and object boundaries rather than raw pixel intensities, making the model more resilient to changes in imaging conditions.

The incorporation of structural priors also plays a crucial role in improving generalization. By guiding the refinement process with contour information, the model enforces shape consistency and reduces the likelihood of anatomically implausible predictions. This is particularly beneficial in challenging cases where boundaries are weak or partially missing. As observed in the qualitative results, the proposed method maintains coherent and anatomically accurate segmentations even under severe noise and low-contrast conditions, whereas baseline methods often produce fragmented or distorted outputs.

Overall, the cross-dataset evaluation confirms that the proposed contour-guided framework effectively addresses the limitations of existing methods in handling domain variability. By integrating structural priors with attention-based refinement, CGQR-Net achieves improved robustness and reliable performance across different datasets, highlighting its potential for real-world clinical applications.

\section{Discussion}
\label{discussion}
The experimental results demonstrate that the proposed CGQR-Net consistently outperforms existing methods across both CAMUS and CardiacNet datasets, indicating strong segmentation capability as well as robust generalization. Unlike conventional approaches that rely primarily on pixel-wise supervision, the proposed method explicitly incorporates structural information through contour-guided query refinement, which proves to be critical for improving both accuracy and boundary delineation.

From the quantitative results, CGQR-Net achieves the highest DSCs across all structures on both datasets, with particularly noticeable improvements in challenging regions such as the epicardium and left atrium. These structures are known to exhibit weak boundaries and high variability in echocardiographic images. The consistent gains across both datasets suggest that the model is not simply overfitting to dataset-specific characteristics but is effectively learning transferable structural representations. This is further supported by the external validation on CardiacNet, where the model maintains superior performance despite significant domain shift.

The qualitative results provide deeper insight into these improvements. Existing methods tend to produce either overly smooth segmentations or fragmented predictions, particularly under low-contrast and noisy conditions. In contrast, CGQR-Net generates sharper and more anatomically consistent boundaries. The contour points extracted from coarse predictions act as explicit structural cues, guiding the refinement process. This allows the model to correct coarse segmentation errors, reduce leakage into surrounding regions, and preserve fine details that are typically lost in standard CNN-based methods. The boundary predictions further confirm that the model learns meaningful edge representations, which directly contribute to improved segmentation quality.

The ablation studies reinforce the importance of each component in the proposed framework. Removing the contour query mechanism results in a noticeable performance drop, particularly in boundary-sensitive regions, highlighting its role in encoding structural priors. Similarly, the absence of the boundary head leads to degraded edge quality, indicating that auxiliary boundary supervision is essential for precise delineation. The feature pyramid fusion and cross-attention modules also contribute significantly by enabling effective interaction between multi-scale features and contour-based queries. These findings confirm that the performance gains are not due to a single component but arise from the synergistic integration of all modules.

From a domain generalization perspective, the proposed method demonstrates clear advantages over existing approaches. Traditional CNN-based models are highly sensitive to variations in intensity distribution and noise patterns, which limits their performance on unseen datasets. In contrast, CGQR-Net leverages contour-based representations that are inherently more invariant to such variations. By focusing on structural information rather than purely intensity-based features, the model achieves better robustness across datasets with different imaging characteristics. This is particularly important in real-world clinical scenarios, where data variability is unavoidable.

Despite these strengths, the proposed method still has limitations. The contour extraction process depends on the quality of the coarse segmentation, and errors at this stage may propagate to the refinement module. In extremely challenging cases with severe noise or missing structures, the contour representation may become less reliable. Additionally, the use of cross-attention introduces additional computational overhead compared to standard CNN models, which may impact deployment in resource-constrained environments.

Overall, the proposed CGQR-Net effectively addresses the key limitations of existing segmentation methods by integrating structural priors, multi-scale feature pyramid fusion, and attention-based refinement. The consistent improvements observed in both quantitative and qualitative evaluations, along with strong generalization performance, demonstrate the potential of contour-guided query learning as a powerful paradigm for medical image segmentation.

\section{Conclusion}
\label{conclusion}
This paper introduced CGQR-Net, a boundary-aware segmentation framework that integrates contour-guided query refinement with multi-scale feature learning for accurate cardiac structure segmentation. The proposed method leverages contour-derived structural priors and cross-attention refinement to enhance boundary delineation and shape consistency. Experimental results show that CGQR-Net achieves superior performance on the CAMUS dataset and maintains strong generalization on the CardiacNet dataset, demonstrating robustness to domain variations and imaging conditions. The findings confirm that boundary-aware refinement plays a crucial role in improving segmentation quality, particularly in challenging regions with weak or ambiguous boundaries. Overall, the proposed approach offers a reliable and effective solution for cardiac image segmentation, with strong potential for clinical application.

\section*{CRediT authorship contribution statement}
\textbf{Zahid Ullah:} Conceptualization, Methodology, Software, Formal analysis, Investigation, Data curation, Writing - original draft, Writing - review \& editing. \textbf{Sieun Choi:} Writing - original draft, Writing - review \& editing. \textbf{Jihie Kim:} Formal analysis, Investigation, Supervision, Project administration, Project management.

\section*{\textbf{Declaration of Competing Interests}} The authors declare that they have no known competing financial interests or personal relationships that could have appeared to influence the work reported in this paper.

% \section*{Acknowledgment}
% {This research was supported by the ``Regional Innovation System \& Education (RISE)'' through the Seoul RISE Center, funded by the MOE (Ministry of Education) and the Seoul Metropolitan Government, and conducted in collaboration with HuVet bio, Inc. (https://www.huvetbio.com/) (2026-RISE-01-007-05), and the Artificial Intelligence Convergence Innovation Human Resources Development supervised by the MSIT (Ministry of Science and ICT) and the IITP (Institute for Information \& Communications Technology Planning \& Evaluation) (IITP-2026-RS-2023-00254592).}

\label{co}
\bibliographystyle{IEEEtran}
\bibliography{sam.bib}
% \bibliography{FER,sam.bib}

\end{document}